[Title Page]

# An Efficient Approach to Learning Chinese Judgment Document Similarity Based on Knowledge Summarization

**Yinglong Ma[1,*], Peng Zhang[1] and Jiangang Ma[2]**

[1]School of Control and Computer Engineering, North China Electric Power University, Beijing 102206, China

[2]Renmin University of China Law School, Beijing 100872, China

\* Correspondence author

**Correspondence information:**

**Full name: Yinglong Ma**

**Affiliation:** School of Control and Computer Engineering,

North China Electric Power University, Beijing 102206, China,

Email address: yinglongma@gmail.com

Telephone number: +86 10 61772643

# An Efficient Approach to Learning Chinese Judgment Document Similarity Based on Knowledge Summarization

**Abstract**

A previous similar case in common law systems can be used as a reference with respect to the current case such that identical situations can be treated similarly in every case. However, current approaches for judgment document similarity computation failed to capture the core semantics of judgment documents and therefore suffer from lower accuracy and higher computation complexity. In this paper, a knowledge block summarization based machine learning approach is proposed to compute the semantic similarity of Chinese judgment documents. By utilizing domain ontologies for judgment documents, the core semantics of Chinese judgment documents is summarized based on knowledge blocks. Then the WMD algorithm is used to calculate the similarity between knowledge blocks. At last, the related experiments were made to illustrate that our approach is very effective and efficient in achieving higher accuracy and faster computation speed in comparison with the traditional approaches.

*Keywords:* Legal document; Domain ontology; Text similarity computation; WMD.

# 1. Introduction

In the last decades, we have witnessed an increasing number of digitally available legal documents with the drastic developments of information technology. Legal document search has received more attention in current legal applications such as case based reasoning [1,2] and legal citations [3], etc. In order to ensure that identical situations can be treated similarly in every case, a previous similar case in common law systems can be used as a reference with respect to the current case. Therefore, one of key issues in judgment document search is to retrieve relevant prior documents with respect to a current document, which is also called precedent retrieval [4,5]. It is necessary for an automatic precedent retrieval system to help legal practitioners to accurately refer to the previous cases that are most similar to the current case.

How to find legal documents that a user needs indeed or that are closely relevant to what a user really wants, has become a crucial problem for effective and efficient legal document search, which heavily depends on the degree to which legal documents can be semantically represented and understood. Currently, understanding and automatic processing of the semantic contents of judgment documents has become a challenging task. On one hand, the activities involving judgments are rather complex and a judgment probably contains different domain specific formats and semantics. In the situation, similar case documents are possibly written in different terminologies even if there does have a common legal dictionary. If a judgment document is not written in formal terminologies, it is not easy to search relevant documents by using indexed keywords. On the other hand, different legal practitioners do not always have unambiguous understanding with respect to a certain case, and therefore they often have different judgments to the case because of judicial discretion. All the aforementioned issues will make it very difficult for document search to unambiguously understand and automatically process the semantics of judgment documents. What is most important is that a very high quality search (e.g., high precision) is urgently needed for judgment document search, unlike widely accessible documents on the Internet.

Most of existing approaches failed to capture semantic contents of judgment documents. The traditional strategy for document search is to index keywords of documents and further rank documents containing keywords according to the relevance to the keywords that users input. Some of existing approaches are to compare two documents by treating them as bag-of-words (BOW) [6]. Each term of a document is given weight according to TF-IDF method. The vector space model (VSM) and its variations [7,8] are also very helpful in computing the similarity between documents. These methods mentioned above are lack of semantic understanding of judgment documents. Currently, some popular approaches, such as information extraction based on rule-based approach [9], and machine learning techniques [10,11,12,34], have been widely used in retrieving the previous similar cases with respect to the current case. Some semantic representations, such as Word2Vec [13], Doc2Vec [14] and Category2Vec [35], etc., as distributive representation models, have become a very attractive approach, which uses neural networks that are trained to reconstruct linguistic contexts of words [29]. The word embedding approach can capture multiple different degrees of similarity between words and allow words with similar meaning to have a similar representation, and therefore can be used for analyzing the latent semantics of unstructured documents [30]. However, it is often difficult to interpret representations learned from data with accurate high-level semantics [31]. Essentially speaking, it often is sensitive to training data distribution, so poorly distributed data can reach an inferior or even wrong generalization.

We argue that formal semantics, the domain knowledge capturing high-level semantics, has undoubtedly provided a way to systematically encode, share, and reuse knowledge across applications and domains, which can support a wide range of key aspects in machine learning, data mining, and artificial intelligence techniques. As for the similarity computation of legal documents, the use of domain knowledge must be helpful in filtering out redundant or inconsistent data, generating semantic enriched results and improving the effectiveness and efficiency of legal document processing.

In this paper, we propose a knowledge block based machine learning approach to computing Chinese judgment document similarity for searching previous similar case judgment. The core semantics of Chinese judgment documents can be summarized by knowledge blocks utilizing domain ontologies for judgment documents. Then the Word Mover's Distance (WMD) algorithm [15] is adapted to calculate the similarity between knowledge blocks. At last, the related experiments were made to illustrate that our approach is very effective and efficient in achieving higher accuracy and faster computation speed in comparison with the traditional approaches.

This paper is organized as follows. Section 2 is the related work. In Section 3, we first discuss the construction of the top ontology and the domain ontologies corresponding to different types of crime, and the judgment document information summarization based on domain knowledge. Section 4 is to discuss the similarity computation of judgment documents based on WMD. We validated the effectiveness and efficiency of our approach by using the k-NN algorithm based judgment document classification.

2. Related Work

With the development of information technology in the judicial field, the number of digital judgment documents has rapidly increased. Text processing and analysis based on Chinese legal documents has attracted more and more attention [33]. Koniaris et al. presented an approach for extracting a machine readable semantic representation from unstructured legal document formats [16]. Wyner applied natural language processing tools to textual elements in legal cases that are unstructured to produce annotated text, from which information can be extracted for further processing and analysis [17]. Zhang et al. proposed a circular ontology between normative documents and judicial cases in order to contribute open-textured legal concepts and improve the retrieval accuracy [18]. Chou et al. developed a document classification, clustering and search methodology based on neural network technology that helps law enforcement department to manage criminal written judgments more efficiently [19]. The processing and analysis based on judgment documents recently has made good progress, which helps judicial practitioners greatly.

The key of understanding and analysing judgment documents is to find a way in which the semantic information of judgment documents can be easily captured and described effectively and efficiently. It is often difficult for traditional document processing to capture and describe the semantic-level features. Most of existing approaches use the popular models such as bag-of-words (BOW) and TF-IDF for document representations [20], and therefore the similarity measures based on both them often rely on computing word overlap. In other words, the similarity between two documents is 0 if they don't have words in common. This is somewhat unreasonable because two documents containing completely different words may also express the same meaning. This issue is mainly caused by the heterogeneity of semantic representation such as synonymy and polysemy. Not relying on word overlap, there are some approaches that can understand documents semantically by using statistical models such as topic model. The two semantic analysis models, Latent Semantic Indexing (LSI) [21] and Latent Dirichlet Allocation (LDA) [22], map the words and documents into a topic vector space to solve the problem of synonymy and polysemy. A document vector is low dimensional and some noise has been removed, which improve the accuracy of documents search and reduce the search time. However, both LSI and LDA are computationally expensive although approaches like LSI and LDA have made great improvement in the last decade.

Many new approaches for text analysis are based on word embeddings [23,24], which is generated by a neural network architecture. The word embedding is a distributive representation of a single word. As opposed to the one-hot vector used in BOW, a distributed vector is dense, low-dimensional and continuous. In a word embedding model, words that have similar meaning have similar representations and therefore word embeddings are suitable for helping in learning the latent semantics of documents. Word embeddings can capture the semantic information of words well, which is leveraged by many approaches. Word Mover's Distance (WMD) is just an example that can compute the semantic similarity between two documents by utilizing word embeddings and Earth Mover's Distance (EMD) model [15]. In comparing with different text classification tasks based on text similarity, WMD was proved to outperform some popular models such as LSI and LDA [15]. However, WMD is computationally expensive for long documents. And also, some document-level semantic information cannot be captured for WMD, instead of the word-level semantic information due to the fact that WMD only takes the word2vec based vector representation as input. The word2vec representation cannot capture the document-level semantics fully. It is often difficult for the word2vec vectors to interpret representations learned from data with accurate high-level semantics (e.g., document-level semantics). Essentially speaking, it often is sensitive to training data distribution, so poorly distributed data can reach an inferior or even wrong generalization.

Semantically understanding the judgment documents is more challenging because judgment documents are often more complicated [25,26] than the open domain documents. We argue that the domain knowledge developed by domain experts can better capture the high-level semantics of legal documents, and has undoubtedly provided a way to systematically encode, share, and reuse knowledge across applications and domains. In addition, the use of domain knowledge must be helpful in filtering out redundant or inconsistent data, generating semantic enriched results and improving the effectiveness and efficiency of legal document processing.

The contributions of this paper are as follows.

First, we propose a knowledge block based machine learning approach to computing Chinese judgment document similarity for searching previous similar case judgments.

Second, the domain ontologies are constructed to guide summarization of the knowledge block of Chinese judgment documents.

Third, the WMD based machine learning approach is adapted to calculate the similarity between their summarized semantic descriptions by using WMD distance. The related experiment results show that our approach is very effective and efficient in achieving higher accuracy and faster computation speed in comparison with the traditional approaches.

## 3. Overview of Approach

In this section, we will give a brief description about our approach to Chinese judgment document similarity computation based on knowledge blocks summarization, which is shown in Figure 1. Strictly speaking, our work can be divided into three parts: domain ontologies for judgment domain knowledge, automatic summarization based on knowledge blocks, and the classification based judgment document similarity evaluation.

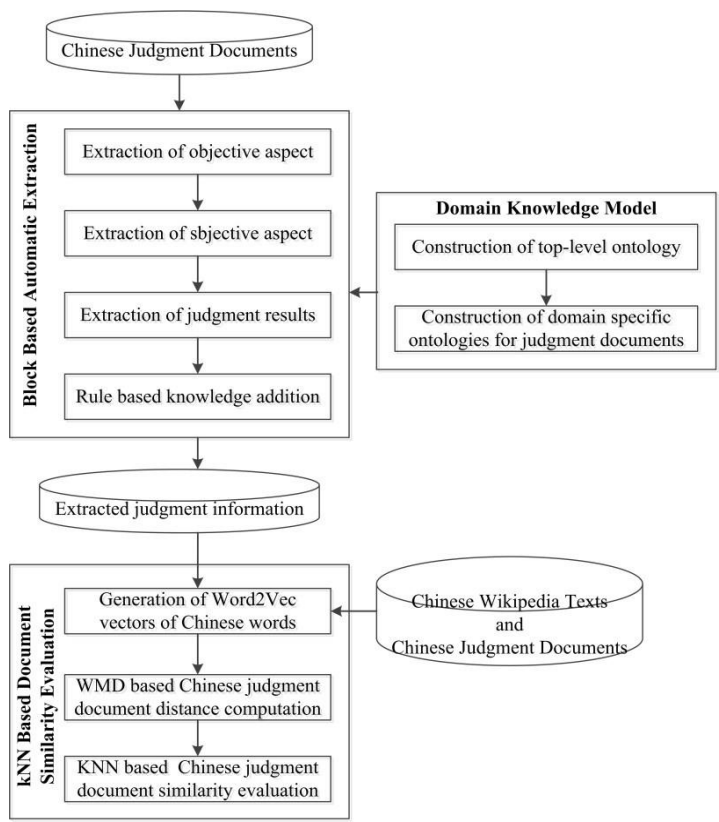

Figure 1.The Overview of our Approach

The domain knowledge for judgment documents can be modeled by the ontology approach that has been widely used in the fields of knowledge representation, artificial intelligence and knowledge engineering [32]. Ontologies provide the formal and shared conceptualization of domain knowledge, and have the expressive ability to explicitly represent classes (concepts) and the relationships between classes such as concept subsumption and semantic association between classes. As for the domain of judgment, judgments often are associated with social situations and activities, which are involved in some terms related to judgment, such as agents, actions, processes, time, space, etc. Two types of ontologies will be constructed. We will use a foundational ontology to model the concepts and relationships of law spreading over almost the full range of common sense, which is called the top-level ontology. The knowledge in the top-level ontology will be very abstract for understanding the relevant concepts in the judgment domain. Furthermore, domain-specific concepts and relationships are modeled by different domain specific ontologies in terms of their judgment results due to the fact that they involve different accusations, sorts of punishment, and criminal details. The concept in a domain-specific ontology can be connected to some concept in the top-level ontology. The combination of the two types of ontologies will be used as the foundation of knowledge summarization and processing of judgment documents.

The second part is the automatic summarization based on knowledge blocks. A knowledge block means a distinguishable aspect of contents contained in judgment documents. According to the writing style and norms of Chinese judgment documents, a Chinese judgment document consists of some distinguishable knowledge blocks such as basic document information, criminal object, criminal subject, objective aspect, subjective aspect, judgment results, etc. Some blocks can be further divided into many fine-grained knowledge fragments that possibly reside in different paragraphs in a document. For judges and other legal practitioners, the core knowledge blocks including objective aspect, subjective aspect and judgment results are very important to help them search previous similar case judgments. So in this part, information summarization technology is used to find the most relevant paragraphs that completely cover the contents of these core knowledge blocks. In addition, rule based approaches are used to add some new knowledge into knowledge blocks according to the China laws in order to qualitatively analyze the summarized contents.

In the third part, we compute the similarity between Chinese judgment documents by using the WMD method. The corpus includes a lot of Chinese judgment documents and the Chinese Wikipedia texts. We use a word segmentation system to segment every Chinese text into a set of Chinese words, and further form the normalized bags of words model for the corpus that is trained to generate a vector for each of Chinese words based on the Word2Vec model. A Chinese judgment document is represented as a set of vectors representing Chinese words that are included in the document. Then the WMD method is to calculate and obtain the distance between two documents. At last, we use the kNN (K-Nearest Neighbours) [28] algorithm to evaluate the effectiveness and efficiency of our approach due to that kNN is one of the most popular classifiers [36].

**4. Domain Knowledge Model for Judgment Documents**

Judgment domain knowledge can be described by ontologies that provide a common vocabulary of a domain of interest, and define the meaning of the terms and the relationships between them [32]. Nowadays, ontologies have been widely applied in many fields such as knowledge engineering and artificial intelligence. Generally speaking, judgment refers to social situations and activities in general terms. It is the nature of these social situations and activities that is the object of ontological modeling of law. For modeling and understanding some judgment domain, some notions about agents, actions, processes, time, space, etc, should be included. On one hand, a foundational ontology for judgment domain appears to be indispensable because the concepts of law are spread over almost the full range of common sense, which should contain our understanding of very abstract concepts, like time, space, causality, physical objects, agenthood, and so forth. We will construct the top-level judgment ontology for describing concepts that are general for all kinds of domains. On the other hand, judgment documents contain many domain-specific concepts due to the fact that they involve different criminal actions, processes, procedures, time, space, persons, roles, intention, etc. The concept in a domain-specific ontology can connect a concept in the top-level ontology. The top-level ontology combined with domain-specific ontologies will be used as the foundation of knowledge sharing for different types of judgment documents.

## 4.1 Top-Level Domain Ontology

The top-level domain ontology describes the common features that all judgment documents have, not caring the exclusive details of any specific crimes. In the top ontology, there contain some general concepts about judgment document contents, such as accusation, criminal object, criminal subject, objective aspect, subjective aspect, sort of punishment and sentence, criminal jurisdiction and basic document information, etc.
The top-level ontology is constructed and shown in Figure 2. The basic document information of a judgment document involves the time and place that the document was made, including the judges. The judgment results include the statements claiming that a defendant is guilty. The concept about criminal jurisdiction is to specify which judicial organ carries out the judgment and judicial actions. The criminal object refers to the social relations that criminal behaviours infringe. The concept about sort of punishment and sentence is to declare which sort of punishment the criminal subject is judged and the term of imprisonment. The objective aspect is to describe defendants and their criminal facts and details throughout a crime case. The subjective aspect is used to describe whether the criminal subject has the intention or negligence.

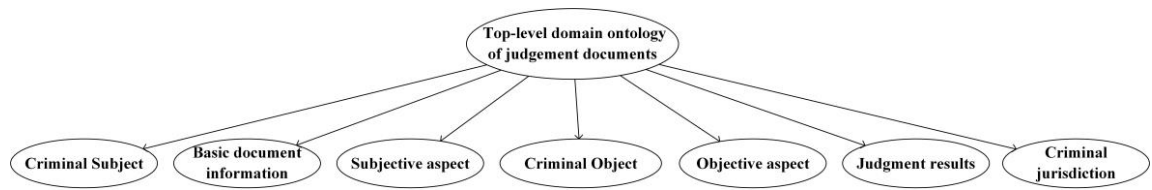

Figure 2. The top-level judgment domain ontology

## 4.2 Domain-Specific Ontology for Judgment Documents

A domain specific ontology for judgment documents is constructed in terms of the accusation and the sorts of crimes contained in judgment documents. The judgment documents with different types of accusations and crimes have a big different in criminal objects, objective and subjective aspects. In a domain specific ontology, much more specific terminological concepts are used to describe the personal records of defendants and their criminal facts and details throughout a crime case. More attention is paid to judging whether the criminal subject obviously has either intention or negligence. These specific terminological concepts will be connected to the corresponding concepts in the top-level ontology, which essentially expands and materialize the terminological concepts in the top-level domain ontology.

We constructed two domain specific ontologies. The one is to describe the domain knowledge for the crime of dangerous driving. The other is about the crime of traffic accident. In the domain ontology for the crime of dangerous driving, the objective aspect can be described in a fine-grained manner. It deals with some representation of subconcepts from different perspectives such as vehicle type, road type, dangerous driving behaviour, etc. Furthermore, each of these subconcepts can be further refined and form more concrete terminological concepts. For examples, road types can be classified into highway, railway, and public parking lot, and so on. Both chasing competing driving and drunken driving belong to the dangerous driving behaviours.

The fragment of the domain ontology for the crime of dangerous driving is shown in Figure 3. What is worthy to note is that we construct the domain knowledge of judgment documents by extending the top-level ontology with newly added and more specific terminological concepts. Due to the lack of space, we only briefly give the expansion in the objective aspect of the domain ontology. A concept in a domain specific ontology can be connected to some concept in the top ontology by using a dashed arrow. The closer to the bottom of the domain ontology, the more specific the terms represent semantics. All of the leaf nodes will have very specific meanings, which characterizes the precise and concrete semantics for judgment documents.

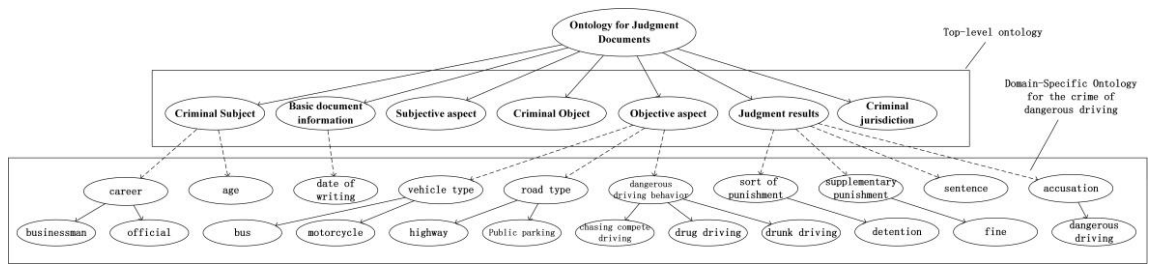

Figure 3. Domain specific ontology for the crime of dangerous driving

We argue that the way in which the domain knowledge of judgment documents can be constructed in an extensible manner [37]. In China laws, there are thousands of accusations and crime types. Each of judgment documents should be involved in one and more accusations and types of crimes, while each of accusations needs to correspond to a domain specific ontology for effective and efficient judgment document processing. On one hand, multiple domain specific ontologies can be seamlessly associated with the top-level ontology for processing and understanding the judgment documents with multiple accusations and types of crimes. For example, there often is a tough association between the crimes of traffic accident and dangerous driving if injured victims (even leading to death) are found in the relevant crime. Assume that a judgment document contains the accusation of both the dangerous driving and the traffic accident, and then the domain ontology of the crime of traffic accident can be integrated into Figure 3 and form new and complex domain knowledge, which is shown in Figure 4. On the other hand, the crimes of traffic accident and dangerous driving have a great similarity due to the fact that both of them contain many common terminological concepts and relations. So the judgment document processing based on the one ontology can be somewhat reused for the other ontology, which will significantly save the cost of domain ontology construction and document processing.

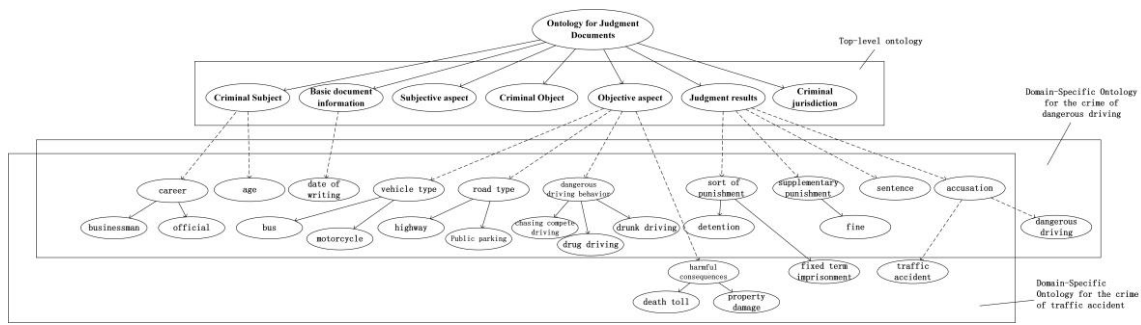

Figure 4. Ontology of documents of traffic accident crime

## 5. Automatic Knowledge Block Summarization

Automatic summarization of Chinese judgment documents is to automatically analyze and obtain the core knowledge for searching the most similar case judgments such that judges should make similar decisions to a case to being tackled by retrieving the most similar case judgments in the historical judgment documents, and eventually achieve the goal that identical situations can be treated similarly in every case.

### 5.1 Automatic Locating and Summarization of Core Knowledge Blocks

As mentioned in the previous section, a judgment document consists of some knowledge blocks that are often characterized by the top-level ontology. However, for analyzing judgment documents, what is important to legal practitioners including judges and lawyers is the objective aspect, subjective aspects and judgment results of a given crime, which are the three core knowledge blocks of Chinese judgment documents. Effective analysis to the three knowledge blocks will make judges have full confidence to make a reasonable decision. Considering the fact, we will make automatic summarization of Chinese judgment documents based on the three knowledge blocks.

The first problem we face is how to locate these core knowledge blocks. A Chinese judgment document naturally has its corresponding domain specific ontologies. The core contents of a document are characterized by the leaf concepts in its corresponding domain specific ontologies. These leaf concepts can be directly or indirectly connected to the concepts in top-level ontology that correspond to the core knowledge blocks. So we locate the information to be summarized by detecting the criminal term strings that can match the leaf concepts in the domain specific ontologies. If a paragraph contains the most criminal term strings can match the leaf concepts w.r.t the core knowledge blocks, then it has the highest possibility to be included in the summarized information. Here, in a domain specific ontology, we denote the finite sets of leaf concepts that correspond to the upper concepts objective aspect, subjective aspects and judgment results as $C_{oa}$, $C_{sa}$ and $C_{jr}$, respectively. For example, for the domain ontology of the crime of dangerous driving in Figure 3, $C_{oa}$={bus, …, public parking lot, chasing compete driving, …, drunk driving}. Assume that a Chinese judgment document D original is denoted as the set D={$P_1, P_2, P_3, \cdots, P_n$}, which consists of its all paragraphs $P_i$ (1≤ i ≤n).

The summarized information of a document is the set of paragraphs that contain the contents of the core knowledge blocks, instead of the set of criminal term strings contained in the document that can match the leaf concepts in the specific ontology. The reason why we do like that is that a simple set of criminal term strings cannot reflect the specific criminal scenarios made by criminal subjects. Algorithm 1 is developed to summarize the relevant knowledge blocks from the given document D. It also can be used for the summarization of a single knowledge block from a document.

In Algorithm 1, $C$ represents the sets of all the leaf concepts related to the core knowledge blocks, $C_{oa}$, $C_{sa}$ and $C_{jr}$ are the sets of leaf concepts about the objective aspect, the subjective aspect and the judgment results. $B_{oa}, B_{sa}, B_{jr}$ are the three summarized knowledge blocks respectively corresponding to $C_{oa}, C_{sa}, C_{jr}$. Variable $count_i$ is used to count the number of the matched term strings relating to the concepts of $C_x$ in the *i*th paragraph.

**Algorithm 1.** Knowledge_Block_Summarization

---

**Input:** A Chinese Judgment Document $D=\{P_1, P_2, P_3, \cdots, P_n\}$, $C=\{C_{oa}, C_{sa}, C_{jr}\}$.

**Output:** The summarized Knowledge blocks $KB=\{B_{oa}, B_{sa}, B_{jr}\}$.

1:   **for** each term set $C_x$ in $C$ **do**
2:       **foreach** paragraph $P_i$ in $D$ **do**
3:          $count_i \leftarrow 0$;
4:          $WordSet_i \leftarrow$ WORD_SEG($P_i$)
5:          **foreach** term string $w$ in $WordSet_i$ **do**
6:             **foreach** concept $t$ in $C_x$ **do**
                //detecting if $w$ and $t$ are matched
7:                 **if** LEV($w,t$)<MAX($|w|,|t|$)/2 **then**
8:                    $count_i$ + +;
9:                 **end if**
10:             **end for**
11:         **end for**
12:      **end for**
13:      $B_x \leftarrow \{P_j | count_j = \text{MAX}(count_1, count_2, count_3, \cdots, count_n)\}$;
14: **end for**
15:    **return** $KB$

---

What is worth to note is there is no separator to separate the Chinese words in a Chinese sentence, unlike English sentences that are separated by the space character. All the Chinese words in a Chinese sentence need to be segmented intelligently. Here, WORD_SEG() is a word segmentation function that is used to cut a Chinese paragraph into a set of Chinese words, which is made in this paper by using the JieBa word segmentation tool [38]. $|w|$ represents the length of string $w$.

The function LEV() is to compute the matching degree between two words by edit distance. Considering the semantic representation of synonymy and polysemy in Chinese words, we use the Levenshtein distance [27] to compute the edit distance between two Chinese strings, which can be mathematically expressed by Formula (1) as follows.

$$lev_{a,b}(i,j) = \begin{cases} \max(i,j) & \text{if } \min(i,j) = 0, \\ \min \begin{cases} lev_{a,b}(i-1,j) + 1 \\ lev_{a,b}(i,j-1) + 1 \\ lev_{a,b}(i-1,j-1) + 1_{(a_i \neq b_j)} \end{cases} & \text{otherwise.} \end{cases} \quad (1)$$

where $lev_{a,b}(i,j)$ is the distance between the first $i$ characters of string $a$ and the first $j$ characters of string $b$. If strings a and b are the same, then their edit distance equals to 0. In Algorithm 1, the function $\text{LEV}(|w|,|t|) = \frac{lev_{w,t}(|w|,|t|)}{\max(|w|,|t|)}$ according to Formula (1). We further set a threshold δ, when $\text{LEV}(|w|,|t|) \leq \delta$, we believe that string *w* can match string *t*.

## 5.2 Rule Based Knowledge Addition

The summarized information, some quantitative information needs to be further tackled. On the one hand, most of existing approaches such as word2vec and WMD is very difficult to capture the semantic features residing in these quantitative numbers although understanding these numbers contained in judgment documents is very important to qualitatively make a decision or judgment. On the other hand, as far as the numbers are concerned, judges often concentrate on the extent of injury rather than the concrete numbers. For example, there are two cases about the crime of traffic accident. In one case, the number of deaths is five, and one person was injured. But in the other case, the crash in the traffic accident killed 1 and injured more than 5. Both cases include the same numbers, but they have very big difference to qualitatively analyze the crime and make a judgment in terms of the extent of injury.

We use rule based approaches to add some new knowledge into knowledge blocks according to the China laws in order to qualitatively analyze the summarized contents. Especially for the knowledge block of objective aspect, the information about deaths and injuries should be added into the block by replacing the numbers of deaths and injuries with a description about the extent of injuries.

By using the regular expression based approach, we first locate and summarize the strings containing the numbers of deaths and injuries, respectively. We can respectively obtain the numbers corresponding to deaths and injuries, through which we can further compute the intervals they belong to. Different intervals mean different levels of the extent of injuries. As an example of a regular expression, the pattern "caused (.*) deaths" works in most cases of finding the death toll, where ".*" are meta-characters that match Chinese numbers related to the death toll. Considering that different legal practitioners possibly have different habits and styles to describe the number of deaths and injuries, we look through and review more than 10,000 Chinese judgment documents from which we summarize all the possible writing patterns describing deaths and injuries and establish a regular expression for each of them to cope with the changeable syntactic structures.

Second, we define a serial of replacement rules with respect to the extent of injury. Specifically speaking, the extent of injury can be classified into different levels according to the intervals of deaths and injuries. For examples, minor injuries and no death belong to the general level of traffic accident, deaths within 1 to 2 persons belong to the serious level, and deaths more than 3 people are the extraordinarily serious level. Different levels will be considered for possibly different sorts of punishment and sentence. The replacement rules for the extent of injury are as follows.

"A general level of casualties" ← deaths [0, 0] ∨ injuries≥0.
"A serious level of casualties" ← deaths [1, 2]
"An extraordinarily serious level of casualties" ← deaths ≥3.

By the replacement rules, the knowledge block will be added about the description of the extent of injuries for automatic analysis and processing of the future.

## 6. WMD Similarity Computation Based on Knowledge Blocks

In this section, we first use WMD approach to calculate the similarity between any two Chinese documents. What is worth to note is that knowledge blocks are used for WMD computation, instead of the original Chinese judgment documents. An important reason is that we want to reduce the time complexity of document similarity computation without loss of accuracy because the WMD approach is rather time consuming when tackle massive data.

In WMD, a text document $\mathbf{d} \in R^n$, is represented as a normalized bag-of-words (**nBOW**) vector, where $n$ represents the dimensions of the nBOW model, and $d_i = \frac{c_i}{\sum_{j=1}^{n} c_j}$ if word $i$ appears $c_i$ times in the document. WMD leverages word2vec model to generate high-quality word embeddings based on large-scale data sets. The semantic similarity between two words can be measured by their Euclidean distance in the word2vec embedding space. To be precise, let $c(i,j) = \left\| x_i - x_j \right\|_2$ be the distance between word $i$ and word $j$, where $x_i$ and $x_j$ are the word vectors of word $i$ and word $j$. $c(i,j)$ is also called as word travel cost.

WMD incorporates **nBOW** vectors with the travel cost to compute the document distance. Assume that two text documents $\mathbf{d}$ and $\mathbf{d}'$ have been represented as **nBOW** vectors, and each word $i$ in $\mathbf{d}$ will be transformed into any word in $\mathbf{d}'$ in total or in parts. A flow matrix $T \in R^{n \times n}$ is used to indicate the specific amount of transfers where $T_{ij}$ denotes how much of word $i$ in $\mathbf{d}$ travels to word $j$ in $\mathbf{d}'$. There are two restrictions: $\sum_j T_{ij} = d_i$ and $\sum_i T_{ij} = d'_j$, which mean that the entire outgoing flow from word $i$ in **d** is equal to $d_i$, and the amount of incoming flow to word $j$ in **d'** is equal to $d'_j$. The above two restrictions are to ensure that document **d** can be completely transformed into document **d'**. Then the document distance is defined as the minimum weighted cumulative cost required to move all words from **d** to **d'**, i.e. $\sum_{i,j=1}^{n} T_{ij} c(i,j)$. Formally, WMD can be formulated as a constrained optimization problem.

$$\min_{T \geq 0} \sum_{i,j=1}^{n} T_{ij} c(i,j)$$

Subject to: $\sum_{j=1}^{n} T_{ij} = d_i \quad \forall i \in \{1, \dots, n\}$ (2)

$\sum_{i=1}^{n} T_{ij} = d'_j \quad \forall j \in \{1, \dots, n\}$.

The above optimization problem is a special case of the EMD that is a well-studied transportation problem for which efficient solvers have been developed.

We utilize a corpus including Chinese judgment documents and the Chinese Wikipedia texts to train and generate the Word2Vec vectors for all Chinese words in the corpus. A word segmentation system called JieBa [38] was beforehand used to segment every Chinese text into a set of Chinese words.

In the following, we made word segmentation for the summarized information of every Chinese judgment document and form a set of Chinese words, where stop words were removed. The set of segmented words is transformed to a set of vectors corresponding to these words. The sets of vectors representing two Chinese judgment documents will be used as the input of WMD algorithm for calculating their similarity distance.

## 7. Experiment and Evaluation

In this paper, we propose the knowledge blocks summarization and WMD based approach for Chinese judgment document similarity computation. In order to evaluate the effectiveness and efficiency of our approach, we use the kNN algorithm to make experiments compared with the existing WMD approach (i.e., the WMD based similarity computation based on the original Chinese judgment documents). Two datasets of Chinese judgment documents were used for experimental evaluation. What is important is that WMD has shown a better performance compared against seven popular baselines such as BOW, TFIDF, BM25 Okapi, LDA, LSI, mSDA, and CCG [15], so these baseline methods except WMD will no longer be compared with our approach.

**7.1 Datasets**

There are two datasets in our experiments: the dataset about the crime of traffic accident (CTA), and the dataset about the crime of dangerous driving (CDD). The CTA and CDD datasets respectively include 615 and 687 Chinese judgment documents. Each of Chinese judgment documents is associated with a real judgment case. Both the two datasets were obtained from the China Judgments Online [39], which is official website maintained by the Supreme People's Court of China.

For each of datasets, Chinese judgment documents were classified according to the accusation, sorts of punishment and sentence in the relevant China laws. Specifically, the CTA dataset is classified into four categories such as detention, fixed-time imprisonment within 3 years, fixed-time imprisonment within 3 to 7 years, and fixed-time imprisonment more than 7 years. The CDD dataset is classified into two categories: detention and fixed-time imprisonment. Based on these categories, every document was labeled by the information about the accusation, sorts of punishment and sentence in the judgment results of this document.

**7.2 Word embedding**

A Chinese word embedding is necessary for our experiment. The *word2vec* word embeddings utilized in our experiment is trained on a corpus consisting of the Chinese Wikipedia corpus and 10000 Chinese judgment documents. The training corpus contains about 460,000 Chinese words and the dimensionality of word embeddings is set 400.

**7.3 Experimental Indexes**

We compare and evaluate our approach with the original WMD approach by three experimental indexes: accuracy, efficiency, and efficacy of knowledge blocks summarization.

1) Accuracy: we define Accuracy as follows.

$$accuracy = \frac{\sum_{i=1}^{m} c_{ii}}{\sum_{k=1}^{m} \sum_{l=1}^{m} c_{kl}}$$ , where *m* refers to the number of categories in the corpus and $c_{kl}$ is the number of documents that actually belong to the *k*-th category but are algorithmically classified into the *l*-th category.

2) Efficiency: it can be evaluated mainly by the average performing time that different approaches perform the document similarity computation in order to validate the efficiency of approaches.

3) Proportion of knowledge blocks summarization: it can be used to evaluate the efficacy of knowledge block summarization, and is defined as follows.

$proportion = \frac{ASWE}{ASW}$, where $ASWE = \frac{\sum_{d \in TC} |SWE_d|}{|TC|}$ is the average number of unique words in the summarized information of the original document in the testing set. $ASW = \frac{\sum_{d \in TC} |SW_d|}{|TC|}$ is the average number of unique words in the original document in the testing set, $TC$ is the testing set of the corpus, $SWE_d$ is the set of segmented words of the summarized information in document *d*, and $SW_d$ is the set of segmented words in document *d*.

### 7.4 KNN based Experiments Evaluation and Analysis

We first compute the similarity between Chinese judgment documents by respectively using our approach and the WMD approach. Document similarity computation in our approach was made based on the summarized information of every Chinese judgment document, while the WMD approach used the original documents for computing similarity. The experiments were made on the computer environment with a 3.6GHz processor and the Python programming language.

Then, we use the KNN classifier [28] to compare and evaluate the effectiveness and efficiency of the two approaches. KNN can classify an object according to the labels of its *k* nearest neighbors in the feature space. In every experiment we split the datasets into the training set and the testing set in the proportion of 4:1 randomly. We repeated the computation 5 times and finally obtained their average values as our experimental results.

#### 7.4.1 Comparison of Accuracy

Accuracy evaluation is to validate the effectiveness of the similarity computation of Chinese judgment documents. The higher an approach has the accuracy, the more effective its similarity computation is. We compare the accuracy of both the approaches, which is shown in Figure 5. It is shown that our approach outperforms the original WMD approach over both the datasets CTA and CDD. For the CDD, the accuracy of our approach reaches 90.3 percent compared with 84.8 of the original WMD approach. Out approach has a 5.5% growth in accuracy Compared with the original WMD approach. For the CTA, our approach is much more effective than the original WMD approach. Our approach has the 92.3% accuracy, and obtains an obvious 9.9% growth in accuracy compared with the original WMD approach. The overall experimental results show that our approach has a higher accuracy than the original WMD approach, and can effectively capture the semantics of Chinese judgment documents.

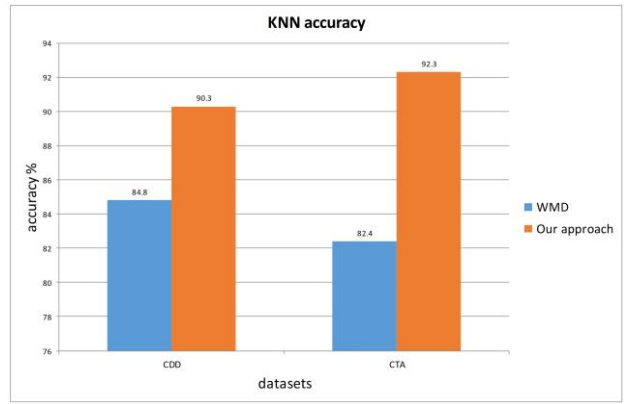

Figure 5. The comparison about accuracy

**7.4.2 Comparison of Efficiency**

In order to validate the efficiency of our approach, we compare the average performing time between both approaches, which is shown in Table 1. For each of approaches, its average performing time (APT) is the addition of the consumed time of word segmentation, document similarity computation, and KNN based classification. There is little difference in the consumed time of word segmentation and KNN classification for both the approaches. The most obvious difference exists in the consumed time of document similarity computation between our approach and the WMD approach. Our approach leads unprecedented low APT in comparison with the original WMD approach. For the CDD dataset, our approach reduces the APT by more than 52 times. The APT is reduced more obviously for the CTA dataset, and our approach is more than 89 times faster than the original WMD approach. So the overall experimental results illustrate that our knowledge block based document similarity computation is very efficient, and has the remarkable improvement in efficiency compared with the original WMD approach.

Table 1. The average performing time on datasets

| Dataset | APT $t_1$ for original documents | APT $t_2$ after knowledge blocks summarization | $t_1/t_2$ |
|---|---|---|---|
| **CDD** | 155190 seconds | 2961 seconds | 52.4 |
| **CTA** | 177704 seconds | 1994 seconds | 89.1 |

**7.4.3 Efficacy of Knowledge Blocks Summarization**

In the followings, we validate the efficacy of knowledge blocks summarization, which is shown in Table 2. It shows the average numbers of unique words respectively before and after blocks summarization for the Chinese judgment documents in both the datasets. It is not difficult for the documents of CDD to find that the document content (average 98 unique words) after block summarization is much less than that (average 278 unique words) before block summarization and the proportion that documents are summarized is about 35.5%. Similarly, for the CTA dataset, the contents of Chinese judgment documents are reduced by more than 30 times.

If we comprehensively analyze the efficacy of knowledge block summarization by combing the experiments about efficiency and accuracy in the previous subsections, it is not difficult to find that our knowledge block summarization is very efficient and effective. In comparison with the original WMD approach based on the same two datasets, our approach has obviously higher accuracy and much faster computation speed by our knowledge bloc summarization. We believe that our knowledge block summarization indeed captures and summarizes the core semantics of these Chinese judgment documents, and therefore has an excellent efficacy for summarizing and capturing the semantics of Chinese judgment documents.

Table 2. The average number of unique words per document

| Dataset | Original documents | After block summarization | Proportion(%) |
|---|---|---|---|
| **CDD** | 278 | 98 | 35.2 |
| **CTA** | 304 | 94 | 30.9 |

**8. Conclusion**

We presented a knowledge block summarization based approach to compute the semantic similarity of Chinese judgment documents for searching the most similar case judgments. We designed and constructed the domain specific ontologies for Chinese judgment documents in terms of different types of crime in an extensible manner. The knowledge block summarization is proposed to summarize the semantic description of each Chinese judgment document. Then, the WMD based similarity computation is made based on knowledge blocks to calculate the similarity between Chinese judgment documents. The related experiments were made to illustrate that our knowledge block summarization based approach are very effective and efficient in achieving a higher accuracy, much faster computation speed.

The future work is to use the deep learning based techniques to find the latent relationships in Chinese judgment documents for predicting the tendency of crime.

**Acknowledgements**

This work is partially supported by the National Key R&D Program of China (2018YFC0830605, 2018YFC0831504) and the National Natural Science Foundation of China (61372182).

**References**
[1] S. Craw, et al. Retrieval, reuse, revision and retention in case-based reasoning. Knowledge Engineering Review, 20(3): 215-240, 2005.
[2] C.Y. Tsai and C.C. Chiu. Developing a Significant Nearest Neighbor Search Method for Effective Case Retrieval in a CBR System. In Proceedings of IACSITSC'09, IEEE, pp.262–266, 2009.
[3] M. Opijnen and C. Santos. On the concept of relevance in legal information retrieval. Artificial Intelligence and Law, 25(1):65-87, 2017.
[4] S. Kumar, et al. Similarity Analysis of Legal Judgments. In Proceedings of COMPUTE 2011, 2011.
[5] D. Thenmozhi, et al. A Text Similarity Approach for Precedence Retrieval from Legal Documents. In Proceedings of Forum for Information Retrieval Evaluation (FIRE), 2017.
[6] Salton, G., Buckley, C.: Term-weighting approaches in automatic text retrieval. Information Processing and Management, 1988.
[7] G. Salton, A. Wong, and C. S. Yang. A vector space model for automatic indexing. Communication on ACM, 18(11):613-620, 1975.
[8] K. Hammouda and M. Kamel. Phrase-based document similarity based on an index graph model. In Proceedings of International Conference on Data Mining (ICDM 2002), IEEE, pp.203- 210, 2002.
[9] F. Galgani, P. Compton, and A. Hoffmann. Lexa: Building knowledge bases for automatic legal citation classification. Expert Systems with Applications, 42(17):6391-6407, 2015.
[10] K. Al-Kofahi, A. Tyrrell, A. Vachher, and P. Jackson. A machine learning approach to prior case retrieval. In Proceedings of the 8th international conference on Artificial intelligence and law (ICAIL 2001), ACM, pp.88-93, 2001.

[11] W. Kim, et al. A Document Query Search Using an Extended Centrality with the Word2vec. In Proceedings of ICEC 2016, 2016.

[12] Q. Lu, W. Keenan, J.G. Conrad, and K. Al-Kofahi. Legal Document Clustering with built in Topic Segmentation, In Proceedings of CIKM 2011, pp.383-392, 2011.

[13] Mikolov, T., Sutskever, I., Chen, K., Corrado, G. S., and Dean, J. Distributed representations of words and phrases and their compositionality. In Proceedings of NIPS, pp. 3111– 3119, 2013.

[14] Q. Le, T. Mikolov. Distributed Represenations of Sentences and Documents. In Proceedings of ICML 2014, 2014.

[15] M. J. Kusner, Y. Sun, N. I. Kolkin, and K. Q. Weinberger. From word embeddings to document distances. In Proceedings of ICML, 2015.

[16] Marios Koniaris, George Papastefanatos, and Yannis Vassiliou. Towards automatic structuring and semantic indexing of legal documents. In Proceedings of the 20[th] Pan-Hellenic Conference on Informatics, 2016.

[17] A. Wyner. Towards annotating and extracting textual legal case elements. Informatica e Diritto: Special Issue on Legal Ontologies and Artificial Intelligent Techniques, 19(1-2):9-18, 2010.

[18] N. Zhang, Y.F. Pu, P. Wang. An Ontology-based Approach for Chinese Legal Information Retrieval. In Proceedings of CENet2015, 2015.

[19] Chou S., Hsing TP. Text Mining Technique for Chinese Written Judgment of Criminal Case. In Proceedings of the 2010 Pacific Asia conference on Intelligence and Security Informatics, 2010.

[20] Salton, G., Buckley, C.: Term-weighting approaches in automatic text retrieval. Information Processing and Management, 24(5):513-523, 1988.

[21] Deerwester, S., Dumais, S.T., Furnas, G.W., Landauer, T.K., Harshman, R.: Indexing by latent semantic analysis. Journal of the American Society for Information Science, 41:391-407, 1990.

[22] Blei, D.M., Ng, A.Y., Jordan, M.I.. Latent dirichlet allocation. Journal of Machine Learning Research. 3: 993-1022, 2003.

[23] T. Kenter and M. de Rijke. Short text similarity with word embeddings. In Proceedings of CIKM, pp.1411–1420. ACM, 2015.

[24] G. Zheng and J. Callan. Learning to reweight terms with distributed representations. In Proceedings of SIGIR, 2015.

[25] G. Boella, et al. Eunomos, a legal document and knowledge management system for the Web to provide relevant, reliable and up-to-date information on the law. Artificial Intelligence and Law, 24(3):245-283, 2016.

[26] R. Nallapati and C.D. Manning. Legal Docket-Entry Classification: Where Machine Learning stumbles. In Proceedings of the 2008 Conference on Empirical Methods in Natural Language Processing (EMNLP), pp.438–446, 2008.

[27] V. I. Levenshtein. Binary codes capable of correcting deletions, insertions, and reversals. Soviet Physics Doklady, 10:707, 1966.

[28] Cover, T. and Hart, P. Nearest neighbor pattern classification. Information Theory, IEEE Transactions on, 13(1): 21-27, 1967.

[29] T. Mikolov, W.T. Yih, and G. Zweig. Linguistic Regularities in Continuous Space Word Representations. In Proceedings of HLT-NAACL, pp.746-751, 2013.

[30] I. Banerjee, M.C. Chen, M.P. Lungren, and D.L. Rubin. Radiology report annotation using intelligent word embeddings: Applied to multi-institutional chest CT cohort. Journal of Biomedical Informatics, 77: 11-20, 2018.

[31] M.D. Zeiler, R. Fergus. Visualizing and understanding convolutional networks. In Proceedings of the 13th European Conference on Computer Vision (ECCV), pp.818-833, 2014.

[32] A. Gomez-Perez, M. Fernandez-lopez and O. Corcho. Ontological Engineering: with Examples from the Areas of Knowledge management, E-commerce and the Semantic Web. Springer, 2004.

[33] Y. Wang, et al. Topic Model Based Text Similarity Measure for Chinese Judgment Document. In Proceedings of ICPCSEE 2017, pp.42-54, 2017.

[34] W. Song, C.H. Li, S.C. Park. Genetic algorithm for text clustering using ontology and evaluating the validity of various semantic similarity measures. Expert Systems with Applications, 36(5):9095-9104, 2009.

[35] G. Zhu and C.A. Iglesias. Exploiting semantic similarity for named entity disambiguation in knowledge graphs. Expert Systems with Applications, 101(1):8-24, 2018.

[36] S. Hussain, J. Keung and A.A. Khan. Software design patterns classification and selection using text categorization approach. Applied Soft Computing, 58:225-244, 2017.

[37] R.C. Chen, Cho-Tscan Bau, Chun-Ju Yeh. Merging domain ontologies based on the WordNet system and Fuzzy Formal Concept Analysis techniques. Applied Soft Computing, 11(2):1908-1923, 2011.

[38] JieBa Word Segmentation. https://github.com/fxsjy/jieba, June 25, 2018.

[39] China Judgments Online. http://wenshu.court.gov.cn/, June 25, 2018.